# Reservoir Computing & Extreme Learning Machines using Pairs of Cellular Automata Rules

Nathan McDonald
Air Force Research Laboratory/ Information Directorate
Rome, NY, USA
Nathan.McDonald.5@us.af.mil

*Abstract*— A framework for implementing reservoir computing (RC) and extreme learning machines (ELMs), two types of artificial neural networks, based on 1D elementary Cellular Automata (CA) is presented, in which two separate CA rules explicitly implement the minimum computational requirements of the reservoir layer: hyperdimensional projection and short-term memory. CAs are cell-based state machines, which evolve in time in accordance with local rules based on a cell's current state and those of its neighbors. Notably, simple single cell shift rules as the memory rule in a fixed edge CA afforded reasonable success in conjunction with a variety of projection rules, potentially significantly reducing the optimal solution search space. Optimal iteration counts for the CA rule pairs can be estimated for some tasks based upon the category of the projection rule. Initial results support future hardware realization, where CAs potentially afford orders of magnitude reduction in size, weight, and power (SWaP) requirements compared with floating point RC implementations.

## I. INTRODUCTION

Reservoir computing (RC) is a relatively recent addition to the field of artificial neural networks (ANN). RC's dynamical behavior makes them well suited to address time-dependent data analysis, which may be found in many machine learning tasks. Unlike typical ANNs which require iterative training for all synaptic connections between all neurons/nodes in the network to be useful, RC works with arbitrarily, sparsely, and statically connected hidden layer neurons called a reservoir [1-2]. Only output neurons' weights are trained to be application specific, and these weights are calculated once via matrix inversion instead of recursive incremental changes. The rest of the neural connections remain static for the duration of the network. The mathematical requirements for this reservoir layer are a) high dimensional projection and b) fading memory [1]. Dynamical systems possessing these characteristics are said to operate at "the edge of chaos." That said, hyperdimensional projection is a powerful computational tool itself and is used by Extreme Learning Machines (ELMs) in a manner similar to RC's reservoir layer but without the short-term memory component [3].

The short list of requirements for a reservoir layer has encouraged research into novel hardware implementations previously unrealistic for other neural network designs, including a bucket of water [4], electronic circuits [5], optics [6, 7], and carbon nanotubes [8]. By exploiting the physics of a hardware reservoir layer itself, the network drastically reduces the many floating point matrix multiplications typically required for ANNs. This makes RC attractive for hardware implementation in size, weight, and power (SWaP) constrained platforms [10].

Interestingly, even networks of Boolean logic gates can demonstrate dynamical behavior. Random Boolean networks (RBNs) are networks of $N$ random Boolean logic functions of $K$ inputs each, allowing for recursive and non-local connections [9]. Though each node $N$ can only have a pair of possible states {0,1}, "edge of chaos" behavior can typically be seen in RBNs of $K = 2$, though such dynamics may be found for other $K$ values [9].

Cellular automata (CA), a special class of RBNs, are attractive as hardware reservoirs because, unlike RBNs generally, CAs follow a homogeneous rule for state transitions based on local interactions between a cell and its immediate neighbors. In particular, one dimensional (1D) CAs, also known as Elementary Cellular Automata (ECA), only have two neighboring cells, the left and right cell, $K = 3$; however, these simple local interactions are sufficient to demonstrate rich dynamical behavior [14, 15].

CAs have only recently been considered for RC and ELMs. A 1D CA based reservoir (ReCA) was first presented in [10-12]. A binary input is randomly projected into a binary vector space and evolved according to a CA rule. The CA state vector is then combined with the next input to create the recurrent connectivity. The entire history of the CA reservoir is used is computing the network output. Important design features included the use of zero buffer vectors to either end of the binary input vector, the use of multiple initial random projections, and the vectorization of the CA reservoir for the purposes of calculating the output weights. Demonstrated applications concerned pathological sequence learning tasks [11] and connectionist-symbolic machine intelligence [10,12], for which the input data were already binarized. Another group proposed a much larger two-dimensional CA space with 256 states for implementing an extreme learning machine (ELM-CA) for edge detection [13].

This work advances the ReCA design concept in several ways. Two CA rules are used in the reservoir layer instead of a single rule, where the reservoir requirements are explicitly implemented with one rule performing hyperdimensional projection and the other rule acting as short-term memory. Whereas previous work focused on cyclical CA edge conditions, by defining the memory rule separately, fixed edge CA reservoirs become a viable option. Further, a unary encoding schemes for real-valued input data is demonstrated in addition to several different reservoir readout schemes.

The rest of the paper is as follows. Section II provides additional background on RC, ELM, and CAs. In Section III, the design of the reservoir's length, depth, and edge properties are discussed and new sequential real-valued input encoding and readout methods are proposed. Section IV presents the experimental results of the ReCA for a variety of benchmarks. In Section V, analysis of results and discussion of future research is provided before final conclusions in Section VI.

## II. BACKGROUND

### A. Reservoir Computing & Extreme Learning Machines

Mathematically speaking, the goal of RC and ELMs is to learn the desired value of an $m$-dimensional output $y(k)$ as a function of $n$-dimensional input $u(k)$, given a number of examples $k = 1, …, K$ (Fig. 1). This is achieved by first projecting the input into a typically larger $N$-dimensional space $x(k)$, called a reservoir in RC. In the particular case of an Echo State Network (ESN) [6], one of two main types of RC network design, for a given input, the reservoir state is

$$x(k) = f(\beta W_{in} u(k) + \alpha W_{res} x(k-1)), \quad (1)$$

where $f$ is the activation function of the reservoir, $\beta$ is the input gain, $\alpha$ is the attenuation of the reservoir state, $W_{in}$ is an $N \times n$ matrix of random projection weights between the neurons from the input layer to the reservoir, and $W_{res}$ is as $N \times N$ matrix describing the internal connections and individual weights of the reservoir neurons. In the case of ELMs, which have no memory component, (1) becomes

$$x(k) = f(\beta W_{in} u(k)). \quad (2)$$

The RC response $\hat{y}(k)$ is the weighted sum of the reservoir output neurons,

$$\hat{y}(k) = W_{out} S(k), \quad (3)$$

where $W_{out}$ is an $N+1 \times m$ matrix of output weights and $S(k)$ is the reservoir state $x(k)$ with the addition of a fixed output bias neuron. Given the target output $y(k)$ for the training examples, $W_{out}$ may be calculated directly by minimizing the Mean Squared Error between the target output $y$ and the reservoir output $S$,

$$W_{out} = (y S^+)^\top, \quad (4)$$

where $S^+$ is the Moore–Penrose pseudoinverse of $S$ and $\top$ denotes the transpose. The most expensive calculation then is the initial matrix inverse calculation as opposed to the iterative gradient descent methods typically employed in ANNs.

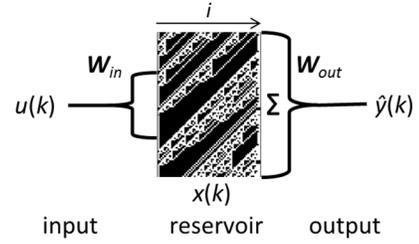

Figure 1: Schematic of CA based reservoir computer (ReCA)

### B. Elementary Cellular Automata

CAs are cell–based state machines which follow a homogeneous rule for state transitions based on local interactions between a cell and its immediate neighbors. 1D ECA are the simplest class of CA, where each cell may only have one of two states {0,1} and whose neighborhood is the cell to its left and right. For every iteration $i$ of the state update rule, a cell's new state is determined with respect to its own state and the state of its two neighbors ($K = 3$) (Fig. 2a). There are then $2^K = 8$ possible 1D neighbor configurations, and $2^{(2^K)}$ Boolean functions of $K$ variables, resulting in 256 unique ECA rules, though only 88 are fundamentally inequivalent [14]. Each ECA rule is numbered according to the decimal equivalent of the rule's output.

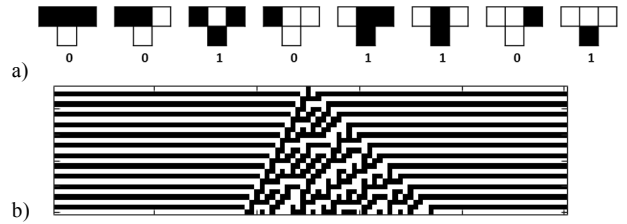

Figure 2: ECA Rule 45 (decimal equivalent to binary 00101101) for a) one iteration of the center cell and b) for 25 iterations of a vector with a single '1' cell in the first row, which generates spontaneous structures in subsequent iterations.

It is the behavior of these rules over multiple iterations that makes CAs useful for reservoir computing (Fig. 2b). CA rules are classified according to four different behaviors: attractor, oscillator, chaotic, and edge of chaos [14], though a rule's category may change over time [15]. A Category I CA's initial response evolves quickly towards a stable, homogeneous state, called an attractor state, e.g. Rule 0, 8, 136 (Fig. 3 a). For Category II CAs, the initial patterns evolve into oscillating or static structures, e.g. Rule 36, 104, 218 (Fig. 3 b).

Additionally, local perturbations tend to remain localized. Conversely, Category III CAs evolve their initial state in a mathematically chaotic manner, spreading out in all available directions and quickly subsuming any apparent structures, e.g. Rule 30, 45, 146 (Fig. 3 c). Category IV CAs operate on the "edge of chaos," demonstrating more mathematically complex behaviors, forming local structures that persist over many iterations, e.g. Rule 110 and 137 (Fig. 3 d). It is these Category IV rules that have been proven to be computationally universal or Turing complete [16]. By combining pairs of rules from different classes, reservoirs of different dynamics can be constructed.

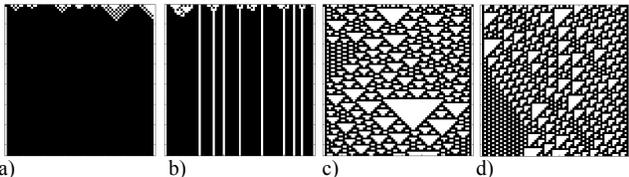

a)   b)   c)   d)

Figure 3: Examples of cellular automata responses to a random initial binary input for rules in a) Category I: Rule 250, b) Category II: Rule 218, c) Category III: Rule 126, and d) Category IV: Rule 110.

III. METHOD

There are a variety of design considerations for implementing a ReCA beyond CA rule selection. This section details these competing design constraints. Since the design of a ReCA for either an RC or an ELM is very similar, the design considerations will be explained with respect to RC with notes about any differences in an ELM design.

*A. Reservoir CA Rule Selection*

For the reservoir to perform useful computation on the input, its response must be repeatable. The choice of a single CA rule for the reservoir layer thus has competing goals: 1) maximize reservoir sensitivity to the input but 2), in the absence of further input, drive the reservoir to a steady state (static or oscillatory). Category III and IV CA rules demonstrate the most input sensitivity but typically do not reach steady state. Category I and II rules have steady states but are often not responsive to the input beyond a handful of iterations.

Instead to achieve both sensitivity and repeatability, two rules are employed (Fig. 4). First, the reservoir is evolved according to a hyperdimensional projection rule nominally chosen from Category III (chaotic) or IV (edge of chaos) for $i_p$ iterations, then it is evolved according to a Category I or II rule for $i_m$ iterations. This second rule, called the memory rule, performs two functions: dimensionality reduction of the CA reservoir state and memory.

The final CA state vector after the memory rule is combined with the next input vector. Since the projection rule is chosen to be very responsive to a sparse input, dimension reduction helps prevent the reservoir dynamics from drowning out the new input. For the ELM implementation, the memory rule may be set to Rule 0 (all cell states become '0') or skipped entirely. Each input $u(k)$ then starts with an unperturbed reservoir.

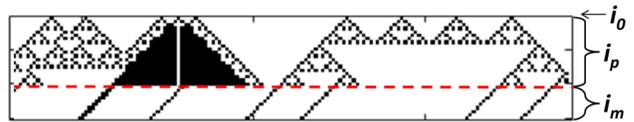

Figure 4: Per input, the reservoir state $x_{i0}(k)$ is evolved according to a projection rule for $i_p$ iterations and a memory rule for $i_m$ iterations. Shown is Rule 218 as the projection rule and Rule 10 as the memory rule.

*B. Reservoir Depth*

The depth of the reservoir is determined by the number of iterations $i$ for each rule. In general, increasing the number of iterations increase the projection dimensionality as the rule response spreads. Beyond a certain $i$ though, the actual input may be drowned out by the dynamics of the rule acting upon itself in the reservoir. Note, if the memory rule is iterated such that it completely suppresses the CA evolution (all cells are in the '0' state), the ReCA behaves as an ELM.

*C. Reservoir Width*

In traditional ANNs, the size of the neural network is simply the number of neurons in the network. For a 1D ReCA implementation, it is primarily a function of the number of columns $j$ in the CA. Since many CA rules cause local perturbations to spread bidirectionality, the binary input vector of length $N$ should be padded on the left and on the right by zero vectors $R$ as buffers, allowing for reasonable information diffusion with time [10]. Thus, the state vector of the reservoir at the $i^{th}$ iteration is of length $L = (N + 2*R)$.

*D. Reservoir Edges*

For CAs, edge cells may be treated in one of two ways: cyclical or fixed. For the cyclical case, an edge cell considers the edge cell on the other extreme as one of its neighbors. This is the typical approach, maximizing information propagation [11]. For fixed edge reservoirs, the edge cells are not evaluated by the CA rule but persist in a fixed state, typically '0'. Their state is used simply to update their only neighboring cell's state in accordance to the given CA rule.

*E. Reservoir Input*

In [11], the binary input is randomly addressed to multiple input vectors, which are evolved concurrently then concatenated before calculating the final reservoir response. In this work, real-valued inputs are encoded into non-random binary vectors. While a decimal to binary number conversion is straightforward to implement, there is an asymmetric change in the binary representation of sequential numbers, e.g. 7 is 0111 but 8 is 1000. That is, a small change in the input caused by noisy data may result in a drastic difference in the ReCA response. Gray coding is similar to binary encoding except that only one element changes in sequential number representations. A third possible encoding scheme is unary encoding, where one element represents one possible input, e.g. 64 decimal places to represent 64 different numbers. While a unary encoder can be inefficient for large input value ranges, it also affords some noise tolerance if the input values are binned to the same element. For simplicity, a unary encoder was used at this time.

Once the input is encoded, two zero vector buffers $R$ are appended to either end of the input vector. The input $u(k+1)$ is then combined via a bitwise XOR operation with the final CA reservoir state vector $x_{ip+im}(k)$ from the previous input (Fig. 5), although any bitwise addition function is expected to suffice [10].

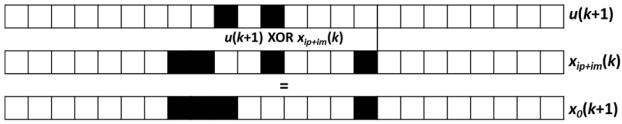

Figure 5: Bit XOR of input $u(k+1)$ with previous ReCA state $x_{ip+im}(k)$

## F. Reservoir Output

After the evolving the reservoir with the two CA rules, there are a few options for reading out the reservoir state $x(k)$. First, for this work, only the $i_p \times L$ reservoir state evolved according to the projection rule is used for the calculation of $\hat{y}(k)$ in (3) (Fig. 6 a). Since the memory rule attenuates the dynamics of the reservoir by design, it is ignored. As with other neural network designs, a bias node is also appended to the reservoir state, $S_k$ is $1 \times L+1$.

### 1) $x_i(k)^{th}$ state

Since the reservoir state at an arbitrary time step or iteration $i$, $x_i(k)$, will be a function of prior inputs, the use of a single row vector is a viable option (Fig 6 b). While very simple to implement, there is an inherent risk that the projection rule may not be sufficiently robust to input noise to be able to attain high accuracy with this binary "all or nothing" state vector.

### 2) Sum of columns

Alternatively, the reservoir state can be represented as the sum of the '1' states per column $j$ over the first $i_p$ iterations of $x(k)$. By normalizing these sums, the node outputs are ensured to be less than 1, which is generally desirable in ANNs. For Category III and IV rules, this approach is particularly attractive, since it may more accurately capture the evolving structure of the CA projection rule in time.

However, if the number of iterations is large or the density of '0' and '1' states is similar, mere sums may not capture the reservoir dynamics over time, since all '1's are treated equally without respect to what iteration they occur at. Smaller bins of iterations may better capture the transient effects of the ReCA response. These bins are then concatenated together (Fig. 6 c), increasing the output node count to $Bj$, where $B$ is the number of bins (Fig. 6 d). The tradeoff is more bins increases the number of nodes but decreases the output resolution of each node.

### 3) Binary/Gray code of column

Finally, the $j^{th}$ column vector may be interpreted as a binary value or Gray code, with the most significant bit being in either the first or the last row (Fig. 6 e). Again the resulting vector values would be normalized.

## IV. EXPERIMENTAL RESULTS

The ReCA framework was applied to several tasks to observe the effectiveness of the approach and gain insight into the effects of the parameter space upon the system. Since the parameter space is quite large, these experiments focused heavily on using shift rules 16 and 2 for the memory layer. For notation, $(a,b)$ indicates Rule $a$ as the projection rule and Rule $b$ as the memory rule. Unless otherwise described, for all experiments, the input was run through an $N = 64$ element unary encoder, two bins were used for the reservoir output, and $R = 64$. The reservoir width was thus $L = N + 2R = 192$.

## A. Sine and Square Wave Classification

The ReCA was first tested on a time-series binary classification problem. In this task, the network must correctly classify the one-dimensional input $u(k)$ as belonging to a sine wave or square wave of amplitude 1 (Fig. 7). The problem is temporally non-linear, since values -1 and 1 could be from either sine or square wave. Random sequences of 200 waves of 20 points each were used for training and testing.

First, using Rule 16 as the memory rule, a search of all 256 ECA rules for the projection rule was performed, where $i_p = 20$ and $i_m = 60$ with fixed edges of '0'. 98 rules achieved 100% accuracy for all 10 trials, with no salient feature in common among these rules except that they included no Category I rules. This experiment was then repeated using cyclical CA edges. This time only 27 rules attained 100% accuracy, and all successful rules were either equilateral or right triangle patterns comprised ≥50% of '1's, e.g. 252, 50, and 182, which are Category I, II, and III rules, respectively. Familiar ECA triangular fractal rules such as 60, 90, and 82 actually obtained among the lowest accuracies, <70%. For comparison, rule pair (110,0), which functions as an ELM, consistently achieved an average accuracy of 95% (routinely misclassifying the '-1' and '1' from the sine wave), which is consistent with a linear classifier tested on an equal number of sine and square waves.

This classification task was performed again for the two other proposed readout schemes: a) the $i^{th}$ iteration, where $i = i_p$, and b) the column binary coding scheme. Using the $i_p^{th}$ iteration, 91 rules attained 100% accuracy. The 7 rules missing from the original 98 successful rules were all left-propagating right-triangles, e.g. rules 70 and 206. However, under the column binary coding scheme, not a single rule attained 100% accuracy. The best accuracy was 97.34%, which was achieved by 172 rules. Since the normalized sum of columns worked best for this simple test case, this was the only readout scheme used for the rest of the benchmark tests.

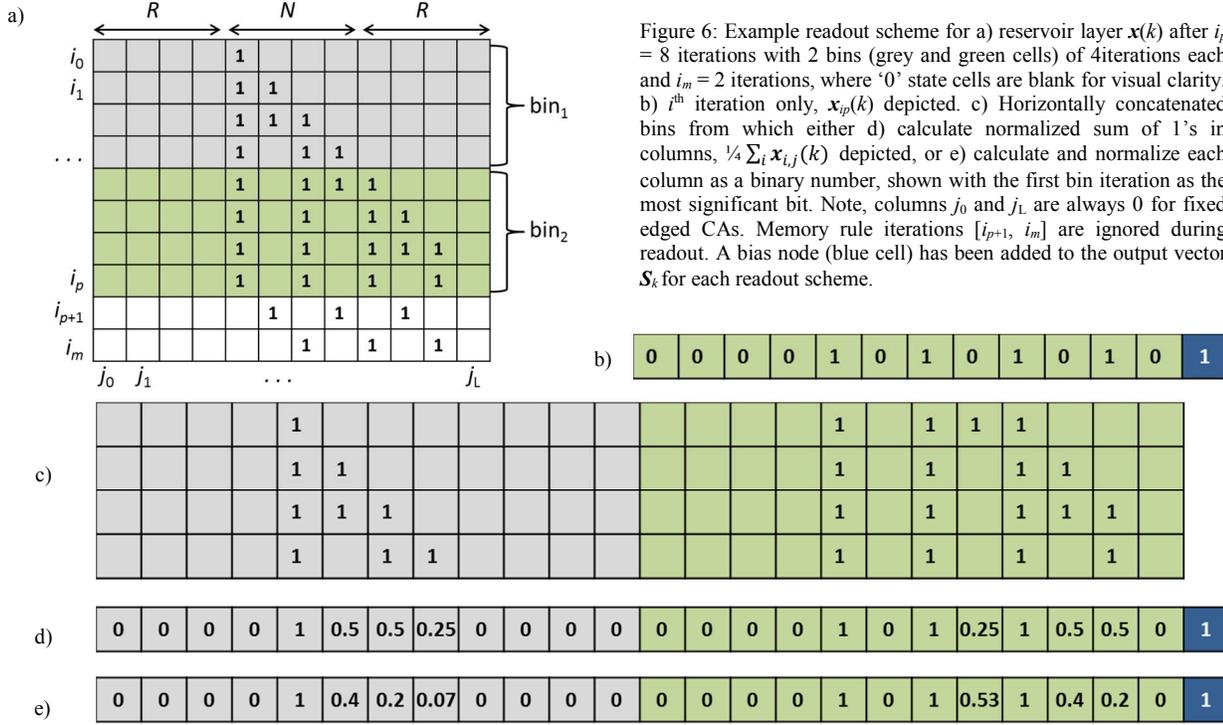

Figure 6: Example readout scheme for a) reservoir layer $x(k)$ after $i_p$ = 8 iterations with 2 bins (grey and green cells) of 4 iterations each and $i_m$ = 2 iterations, where '0' state cells are blank for visual clarity. b) $i^{th}$ iteration only, $x_{ip}(k)$ depicted. c) Horizontally concatenated bins from which either d) calculate normalized sum of 1's in columns, $\frac{1}{4}\sum_i x_{i,j}(k)$ depicted, or e) calculate and normalize each column as a binary number, shown with the first bin iteration as the most significant bit. Note, columns $j_0$ and $j_L$ are always 0 for fixed edged CAs. Memory rule iterations $[i_{p+1}, i_m]$ are ignored during readout. A bias node (blue cell) has been added to the output vector $S_k$ for each readout scheme.

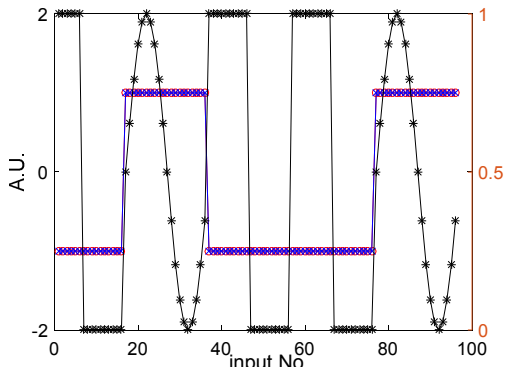

Figure 7: Sine and square wave classification task. The ReCA (blue) correctly classifies the input signal (black) with 100% accuracy (red).

*B. Non-linear Channel Equalization*

One of the standard reservoir computing benchmarks is non-linear channel equalization [6], where a wireless communication channel input signal $d(n)$ is modelled as traveling through multiple paths to a nonlinear and noisy receiver. The task is to reconstruct the original input $d(n)$ from the output $u(n)$ of the receiver. Task performance is measured using Symbol Error Rate. An exhaustive search over the 256 ECA rules for $i_p$ = 20 was performed for both Rule 16 (right shift) and Rule 2 (left shift) for $i_m$ = 60. Each pair of CA rules was run 5 times with 1,000 training points and 100 testing points at 28 dB signal-to-noise ratio (SNR).

In the Rule 16 experiment, 28 rules were able to attain an SER < 0.04 ± 0.02; in the Rule 2 experiment, only 17 rules performed as well, of which 7 were duplicates from the Rule 16 experiment. The best rule pairs were (137,16) at SER = 0.026 ± 0.016, and (242,2) at SER = 0.026 ± 0.019. All 4 CA categories were represented among the diverse successful projection rules, of which Rules 137 and 242 are Category IV and II, respectively. Increasing the training size from 1,000 points to 100,000 points did not statistically improve test accuracy.

The relationship between $i_p$ and $i_m$ was explored for the top two performing projection rules from the Rule 16 experiment: Rules 137 (Class IV) and 36 (Class II). $i_c$ was varied [10, 100] for Rule 16 as $i_p$ varied [10, 100] for both Rule 137 (Fig. 8 a) and Rule 36 (Fig. 8 b).

Since many ECA rules are variants of each other, the sensitivity of the memory rule choice was tested. Nineteen rules visually similar to Rule 16, majority '0' right shift, were used as the memory rule ($i_m$ = 40 – 100) with Rule 137 ($i_p$ = 20) (Fig. 9).

*C. Santa Fe Laser Data*

Another classic benchmark is the Santa Fe laser dataset, where the network must predict a chaotic laser power sequence. An exhaustive search over the 256 ECA rules for $i_p = 20$ was performed for Rule 16 and $i_m = 60$. For this task, a low Normalized Mean Squared Error (NMSE) is desired,

$$NMSE = \frac{\sum_i (y_i - \hat{y}_i)^2}{K \sum_i (y_i)^2}, \quad (5)$$

where $y$ is the target output, $\hat{y}$ is the actual RC output from (3), and $K$ is the length of vector $y$. Since the input data was already scaled [0, 255] a unary encoder of $N = 256$ cells was used. 1,000 training points were used with 200 testing points. An NMSE of 0.0321 was attained with Rule 16 (Fig. 10) as both the projection and memory rule followed by fifteen similar rules in its family (noted in Fig. 9). The next seven best performing rules were checkered equilateral triangles, followed by five more line or triangle rules before other types of rules made the list.

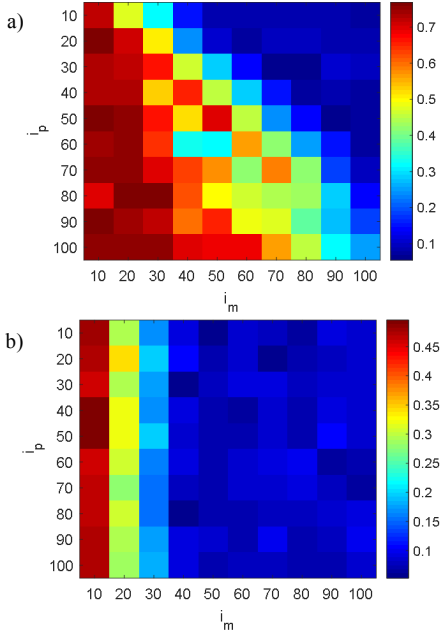

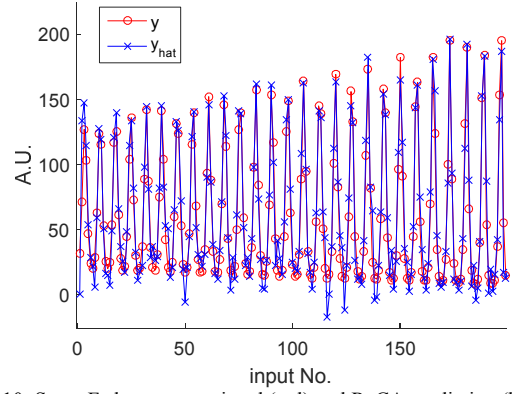

Figure 10: Santa Fe laser target signal (red) and ReCA prediction (blue).

Figure 8: Heatmap of non-linear channel equalization task SER for rule pairs a) (137,16) and b) (36,16) over the range of $i_p$ and $i_m$.

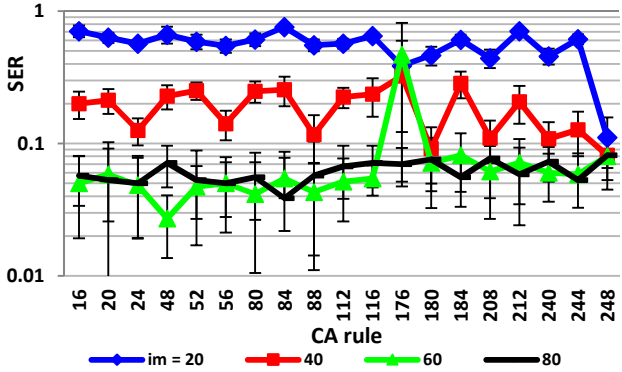

Figure 9: SER as a function of right-shift CA rules as the memory rule when used with Rule 137 for the non-linear channel equalization task.

### D. Iris Classification

The final problem set explores the computational power of CA rules for memoryless random projection networks like ELMs. The iris classification data set is comprised of 3 species of iris (50 examples each) described by 4 attributes [17], where classification accuracies are typically around 95% on account of the small number of examples. Each attribute was converted to a unary value, and each binary attribute vector was concatenated to form one input 147 cells long. 75% (112 samples) were used for training and 25% (38 samples) were used for testing. Since only one CA rule is used for an ELM, all 256 rules at $i_p = 40$ were tested for 50 trials each. While all but a couple rules achieved 100% training accuracies, only 22 rules could generalize the problem to achieve testing classification accuracies greater than 90%. Rule 158 achieved the highest accuracy with 96.2 ± 3.0%, followed by 41 other equilateral triangle rules. For the top five rules, $i_p$ was varied 20 – 80 iterations (Fig. 11), for which there was no statistical difference in accuracies for $i_p > 30$.

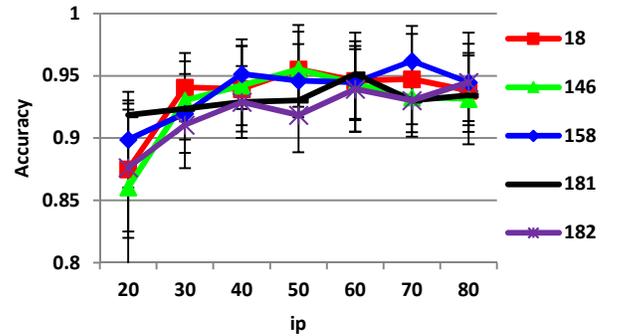

Figure 11: Iris classification accuracy as a function of iteration count $i_p$.

## V. DISCUSSION

Given the thousands of possible ECA rule pairs and design parameter combinations, it is untenable to perform an exhaustive search of the ReCA parameter space. While not rigorous, random rule pair selection was initially attempted for each task, but very few useful rule pairs were produced. By focusing on using CA shift rules for the memory rule across several tasks, some design guidelines are suggested to aid in crafting a particular ReCA solution.

First, while the simplest left (Rule 2) and right shift (Rule 16) rules only shift a single '1' cell one cell over per iteration, when combined with the fixed edges of the CA reservoir, short-term memory is enforced, as the '1' state is lost once it reaches the edge. The degree of influence of prior input, that is the duration of the memory window, can be adjusted in two ways. Increasing the zero buffer length $R$ increases the distance between the input and the reservoir edges, prolonging the number of iterations a signal is active in the reservoir. Alternatively, the number of memory rule iterations can be decreased. Since the last state vector of the CA reservoir is XORed with the next input $u(k)$, the fewer iterations information is laterally displaced, the more iterations it will be evolved according to the projection rule.

At the same time, because the shift rules propagate information away from the length $N$ input vector, new input is less likely to be lost due to the XOR operation. In [10], information preservation was performed by expanding and cyclically shifting the input vector every $u(k)$. Lastly, the interplay between the projection and memory rules may be viewed as a form of information compression and restoration. While shift rules drastically reduce the reservoir state of a Category III or IV rule to a couple of single '1' cells per iteration, these same higher category rules can expand these '1' states sufficiently to achieve the necessary memory effect.

For the sine and square wave classification problem, nearly 40% of the 256 ECA rules achieved 100% accuracy with fixed edges. Category I rules tend to attain uniformity, all '0's or all '1's, within 20 iterations, so it is not surprising that no Category I rules succeeded. When the CA edges were cyclic, the reservoir did not enforce information loss. Since this task requires fading memory for correct classification, the only rules that succeeded enforced their own short-term memory. Rule 16 propagates only edge case '1' states right (Fig. 12 a), so when triangles merge, they remove information from the system (Fig. 12 b). It is expected that fixed edge CA reservoirs will afford more viable projection rule candidates than cyclic edges.

Of the three ReCA readout schemes, the normalized sum of columns consistently afforded the highest accuracy across a majority of the rules. Both the binary readout and the $i^{th}$ iteration readout are very sensitive to a particular cell's state in time, likely making it difficult for the reservoir to generalize a problem.

For the non-linear channel equalization task, the relationship between the projection rule category and the $i_m$ iteration count was explored. For "edge of chaos" Class IV Rule 137, $i_p$ as low as 10 was sufficient to achieve an SER < 0.09 when $i_m \geq 50$. As $i_p$ increased, the minimum $i_m$ for equivalent accuracy also increased by a factor of about 2.

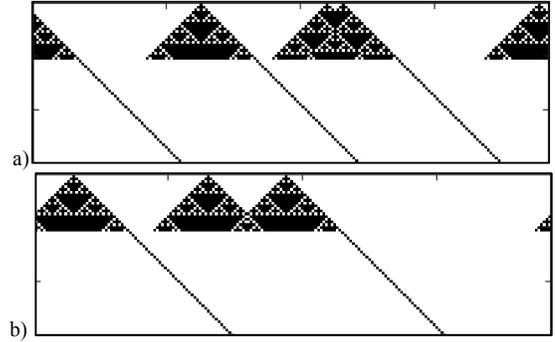

Figure 12: When edges are cyclic, rule pair (182,16) still enforces short-term memory. a) Each isolated triangle response is reduced to a '1' cell shifting right. b) When triangles merge, only the rightmost triangle is encoded by the memory rule.

However, for oscillator Class II Rule 36, the SER was simply dependent upon $i_m$, where $i_m \geq 40$ was required for a similarly low SER. When this problem was performed again using Rule 137 and a collection of memory rules similar to Rule 16, essentially all of the similar rules produced equivalent results when $i_p \geq 60$, consistent with the minimum $i_m$ suggested in Fig. 8. This similarity of behavior is also advantageous from a hardware design perspective, since some ECA rules are certainly more efficient to implement than others, providing some degree of design flexibility.

The best projection rules for the Santa Fe laser data included shift rules (even Rule 16 itself) and checkered triangle rules. Given the normalized sum of columns for the readout scheme, it seems that the position of the unary input was the key piece of information in solving this task. Rules that unambiguously transmit that information performed the best. Further work is required to fully account for this.

The iris data set demonstrated the utility of ECAs for ELM as well. Again, triangle rules performed very well. In particular, equilateral triangles maximized the interactions between all four attributes in time. Since at most an ECA rule can influence $2i-1$ cells per iteration $i$ and since the input vector was 147 elements long, $i_p = 40$ would be sufficiently long to allow a hypothetical input at both ends of the input vector to interact at the center by the $i_p^{th}$ iteration.

While the ReCA demonstrated excellent results for the sine and square wave classification and iris classification tasks, it did not attain state of the art performance on the other RC benchmark tests non-linear channel equalization and Santa Fe laser prediction. A modest ESN typically sees an SER ~1e-4 for the equalization task at 28 SNR [6] and an NMSE of < 0.023 for predicting the laser data [5]. Still the general results using the shift rules for the memory rule are encouraging enough to warrant further investigation into the general ReCA approach.

One significant reason for pursuing a CA based reservoir is the ready means to implement these rules in digital logic gates, greatly reducing the hardware size and computational complexity of the algorithm [18]. Future research will consider a hardware implementation of this ReCA design and a resources analysis will be pursued. The use of a unary encoder was successful across the range of benchmarks, but the described alternative encoding schemes should also be explored. Since there is a wealth of mathematical research describing ECAs and their properties, a more rigorous mathematical analysis of relevant ECA rule characteristics with respect to particular task categories will also be pursued.

## VI. CONCLUSION

A framework for implementing reservoir computing based on 1D cellular automata (ReCA) is presented, in which the hyperdimensional projection and short-term memory of the reservoir layer are segregated and explicitly implemented by two CA rules. Further, left and right shift CA rules in a fixed edge CA space afford reasonable success for a variety of projection rules, potentially significantly reducing the solution search space. Optimal iteration counts for the ECA rule pairs can be estimated for some tasks based upon the category of the projection rule. Future research will explore hardware realization, where CAs potentially afford orders of magnitude reduction in size, weight, and power (SWaP) requirements compared with floating point RC implementations.

## ACKNOWLEDGMENT

The author thanks Dr. Bryant Wysocki for his thoughtful questions and insights. This research was funded by the US Air Force Office of Scientific Research LRIR 15RICOR122. Any opinions, findings and conclusions or recommendations expressed in this material are those of the author and do not necessarily reflect the views of AFRL.